\PassOptionsToPackage{prologue, dvipsnames,table,xcdraw}{xcolor}

\documentclass[sigconf]{acmart}
\AtBeginDocument{%
  }

\copyrightyear{2026}
\acmYear{2026}
\setcopyright{cc}
\setcctype{by}
\acmConference[KDD 2026] {Proceedings of the 32nd ACM SIGKDD Conference on Knowledge Discovery and Data Mining V.1}{August 9--13, 2025}{Jeju Island, Republic of Korea.}
\acmBooktitle{Proceedings of the 32nd ACM SIGKDD Conference on Knowledge Discovery and Data Mining V.1 (KDD 2026), August 9--13, 2025, Jeju Island, Republic of Korea}
\acmISBN{979-8-4007-2258-5/2026/08}
\acmDOI{10.1145/3770854.3780226}
\newcommand{\cheng}{{}}
\usepackage{tcolorbox}
\usepackage{colortbl}
\usepackage{tikz}
\usepackage{enumitem}
\usepackage{multirow}
\usepackage[normalem]{ulem}
\usepackage{graphicx}
\usepackage[table,xcdraw]{xcolor}
\usepackage{threeparttable}
\usepackage{subfig}
\usepackage{algorithm}
\usepackage{algorithmic}
\usepackage{float}
\useunder{\uline}{\ul}{}
\begin{document}
\renewcommand{\algorithmicrequire}{\textbf{Input:}}
\renewcommand{\algorithmicensure}{\textbf{Output:}}
\renewcommand{\algorithmiccomment}[1]{\hfill $\triangleright$ #1}
\title[Wukong Framework for Not Safe For Work Detection in Text-to-Image systems]{{\includegraphics[width=0.7cm]{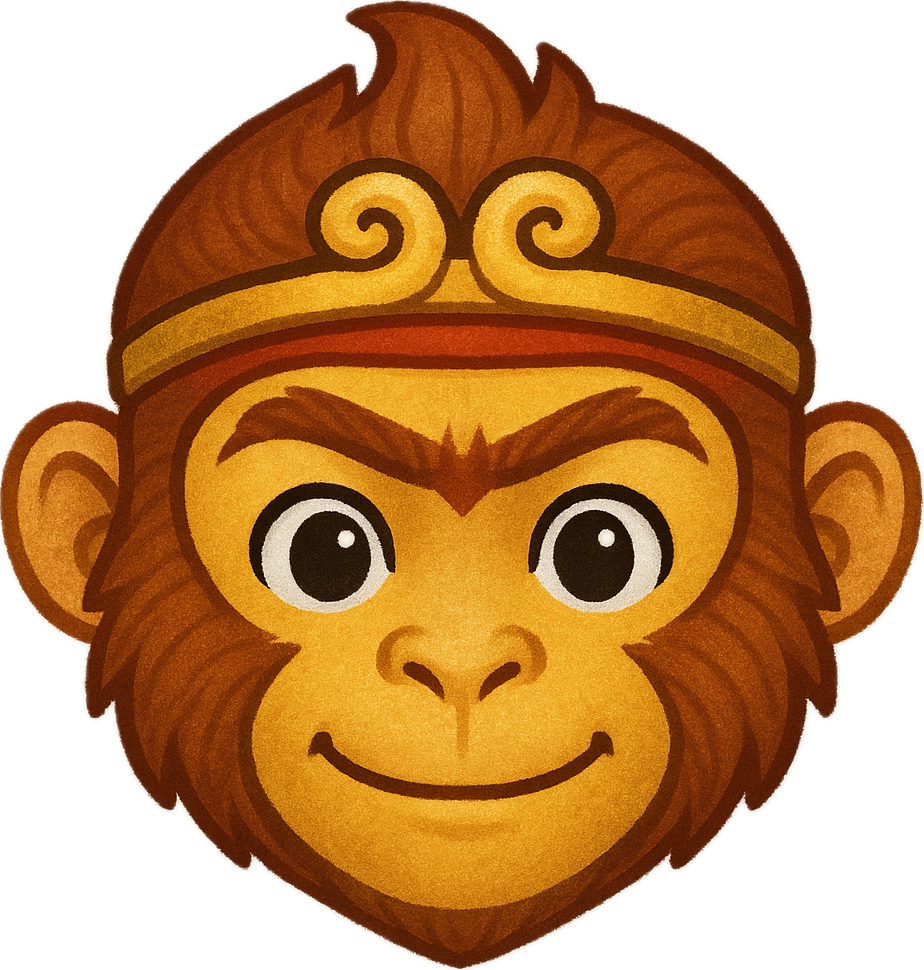}} Wukong Framework for Not Safe For Work Detection in Text-to-Image systems}


\author{Mingrui Liu}
\affiliation{%
  \institution{Nanyang Technological University}
  \city{Singapore}
  \country{Singapore}}
\email{mingrui001@e.ntu.edu.sg}

\author{Sixiao Zhang}
\affiliation{%
  \institution{Nanyang Technological University}
  \city{Singapore}
  \country{Singapore}}
\email{sixiao001@e.ntu.edu.sg}

\author{Cheng Long}
\authornote{Cheng Long is the corresponding author.}
\affiliation{%
  \institution{Nanyang Technological University}
  \city{Singapore}
  \country{Singapore}}
\email{c.long@ntu.edu.sg}







\renewcommand{\shortauthors}{Mingrui Liu, Sixiao Zhang, \& Cheng Long}

\begin{abstract}
Text-to-Image (T2I) generation is a popular AI-generated content (AIGC) technology enabling diverse and creative image synthesis. However, some outputs may contain Not Safe For Work (NSFW) content (e.g., violence), violating community guidelines. Detecting NSFW content efficiently and accurately, known as external safeguarding, is essential. Existing external safeguards fall into two types: text filters, which analyze user prompts but overlook T2I model-specific variations and are prone to adversarial attacks; and image filters, which analyze final generated images but are computationally costly and introduce latency.
Diffusion models, the foundation of modern T2I systems like Stable Diffusion, generate images through iterative denoising using a U-Net architecture with ResNet and Transformer blocks. We observe that: (1) early denoising steps define the semantic layout of the image, and (2) cross-attention layers in U-Net are crucial for aligning text and image regions.
Based on these insights, we propose Wukong, a transformer-based NSFW detection framework that leverages intermediate outputs from early denoising steps and reuses U-Net’s pre-trained cross-attention parameters. Wukong operates within the diffusion process, enabling early detection without waiting for full image generation.
We also introduce a new dataset containing prompts, seeds, and image-specific NSFW labels, and evaluate Wukong on this and two public benchmarks. Results show that Wukong significantly outperforms text-based safeguards and achieves comparable accuracy of image filters, while offering much greater efficiency.
\\{\color{red} \textbf{Warning}: This paper contains potentially offensive text and images.}
\end{abstract}

\ccsdesc[500]{Computing methodologies~Computer vision tasks}
\ccsdesc[500]{Security and privacy~Software and application security}

\keywords{AI security; Text-to-Image generation; Not Safe For Work detection}


\maketitle

\section{Introduction}
Text-to-Image (T2I) generation is a widely used AI-generated content (AIGC) system, enabling diverse and creative image production. However, some generated images may violate community guidelines by including content that is Not Safe For Work (NSFW), such as depictions of explicit nudity (e.g., “a naked woman”). Effectively and efficiently detecting NSFW content in T2I scenarios is essential to ensure these models adhere to ethical standards and community rules.

\begin{figure}[htb]
  \centering
  \includegraphics[width=\linewidth]{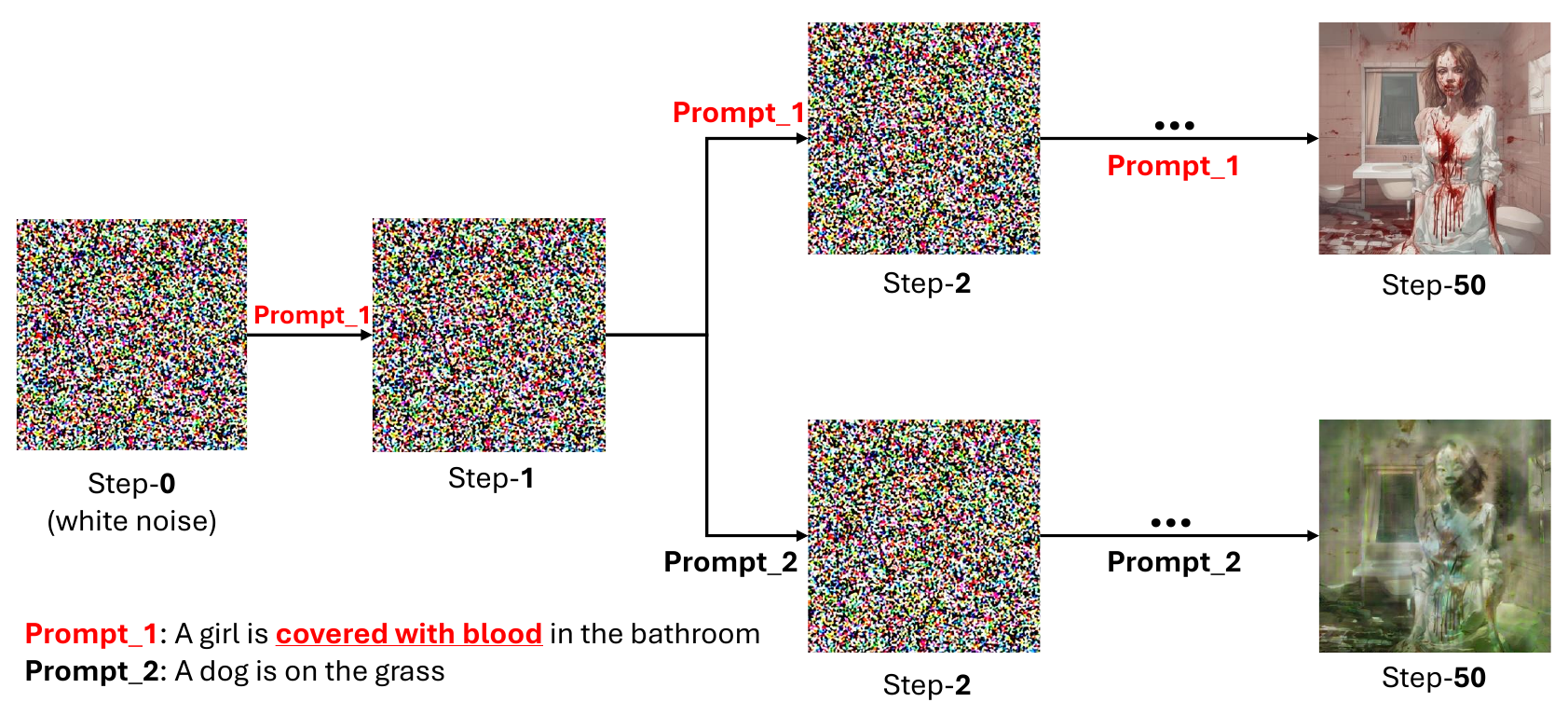}
  \caption{An illustrative example of modifying the textual condition during the early denoising steps in the Stable Diffusion process. The initial latent noise is conditioned with an unsafe prompt (\texttt{prompt\_1}) in the first step. From the second step onward, the first row remains conditioned on \texttt{prompt\_1}, while the second row switches to a safe prompt (\texttt{prompt\_2}).}
  \label{fig: denoising}
\end{figure}

Existing T2I systems have implemented various safety-oriented strategies to prevent the generation of offensive content in images. These defensive methods can be broadly categorized into two types: internal safeguards and external safeguards~\cite{T2Isurvey1}.
Internal safeguards typically involve model-editing techniques aimed at removing unsafe concepts from T2I models (e.g., making the model "forget" the concept of nudity)~\cite{kumari2023ablating, ni2023degeneration, heng2024selective, gandikota2023erasing, orgad2023editing, kim2023towards, wu2024unlearning, kim2024race, huang2024receler, zhang2024defensive}. However, internal safeguards have three main drawbacks: (1) They may degrade image quality in normal (safe) use cases. (2) They remain vulnerable to adversarial attacks (e.g., prompts that appear safe but ultimately generate unsafe images)~\cite{ringabell}. (3) They introduce ethical concerns by misinterpreting user input. 
Additionally, some commercial T2I service providers (e.g., DALL·E 3~\cite{DALLE3}) offer tiered generation services, where regular users can generate only safe images, while subscribed users have access to NSFW image generation. In contrast, internal safeguards cannot provide this functionality, as they remove all unsafe concepts.
Due to these limitations, we focus solely on external safeguards, where the users' generation requests will be rejected if unsafe contents are detected. 
External safeguards can be categorized into text and image filters~\cite{T2Isurvey1}. 
Text filters are simple and efficient but fail to account for T2I model-specific variations, as safety outcomes can vary across models or even within the same model under different random seeds. For example, some textually offensive prompts may generate safe images in certain models, while seemingly harmless prompts can produce unsafe images under specific seeds~\cite{ART, du2023stable, gao2023evaluating, zhuang2023pilot}. Additionally, adversarial techniques~\cite{mma, sneakyprompt} have been developed to bypass text filters, leading to suboptimal performance.
Image filters adopt a post-hoc approach, where images are first generated by T2I models and then evaluated by an image classification model to assess their safety~\cite{Q16, CLIPNSFW, NSFW-Detection-DLr}. While effective, this method incurs high computational costs and significant response delays, as it requires waiting for the full image generation before analysis.
Therefore, there remains a need for an \textbf{external safeguard that can effectively and efficiently determine whether a user's input will generate unsafe images}.

The dominant approach to Text-to-Image generation is the Diffusion series of models~\cite{diffusion}. One typical model, Stable Diffusion~\cite{stablediffusion} begins by initializing a Gaussian noise tensor, followed by iterative denoising using a U-Net architecture that incorporates textual information as guidance, predicting and removing noise at each step.
We visualize the denoising results at each step and empirically observe that the \textbf{initial denoising steps play a crucial role in shaping the semantic content of the generated images}. As shown in Figure~\ref{fig: denoising}, the initial latent noise tensor is conditioned on the unsafe prompt, "\textit{A girl is \underline{covered with blood} in the bathroom}", in the first step. However, even when the prompt is modified to a completely unrelated and safe content, "\textit{A dog is on the grass}", from the second step onward, the generated image still retains unsafe elements (e.g., blood), as seen in the second row of Figure~\ref{fig: denoising}.
This suggests that, despite appearing as white noise to the human eye, the \textbf{early denoising steps have already established the fundamental structure and layout of the image}.
Additionally, we also observe that \textbf{cross-attention layers in U-Net play a key role in mapping textual concepts to image regions}. 

Based on these insights, we propose Wukong, a transformer-based NSFW detector that utilizes intermediate latent representations from early denoising steps as input, {\cheng i.e., it does not need to wait for the whole image diffusion process to be completed so as to improve the efficiency.}
{\cheng It also} leverages pre-trained cross-attention layer parameters in U-Net for feature extraction.

To summarize, our contributions in this paper are as follows:
\begin{itemize}[left=5pt]
    \item We propose Wukong framework for NSFW contents detection. To the best of our knowledge, Wukong is the first external safeguard to leverage intermediate U-Net outputs for NSFW detection, enabling T2I model-specific filtering.
    \item We construct a new dataset for NSFW detection, where each sample includes a textual prompt, a generator seed (ensuring reproducibility of T2I outputs, with each prompt assigned multiple seeds to generate diverse images), and NSFW category-specific labels, obtained using high-quality vision-language models.
    \item We evaluate Wukong on both our newly proposed dataset and two public benchmarks, demonstrating that it outperforms text-based safeguards and achieves comparable accuracy to image-based safeguards, while being significantly more efficient. Additionally, we test Wukong on adversarial (or jailbreaking) prompts, specifically crafted to bypass safeguards by substituting explicitly unsafe terms. The experimental results show that Wukong remains robust against such adversarial prompts. 
    And our code is available\footnote{https://github.com/MINGRUI001/Wukong}.
\end{itemize}


\section{Related Work}
\subsection{Text-to-Image Generation}
Diffusion model~\cite{diffusion} is the dominant architecture in Text-to-Image generation and is widely adopted by models such as Stable Diffusion, DALL·E 3, and Imagen. The mainstream approach involves integrating textual information into diffusion models using a transformer-based architecture, typically a U-Net.

For instance, the Stable Diffusion workflow consists of four main steps:
(1) Text Encoding: The user provides a text prompt, which is converted into textual representation vectors using a text encoder (e.g., CLIP~\cite{CLIP}). (2) Latent Noise Initialization: A random latent noise tensor is initialized, resembling white noise. (3) Iterative Denoising: The model performs a series of iterative denoising steps (typically 50), using the textual representation vectors as conditional inputs. A U-Net-based neural network~\cite{CNNUnet} with a scheduler (such as DDIM~\cite{DDIM} or PNDM) predicts and removes noise at each step, progressively refining the latent representation. (4) Image Decoding: The final latent tensor is decoded using a Variational Autoencoder (VAE) to produce a high-resolution image.


\subsection{External safeguards for T2I models}

Existing external safeguards often rely on detecting forbidden words (i.e., rejecting image generation if the prompt contains terms like "nude"). While this approach is simple and cost-effective, it {\cheng could be} easily bypassed, as malicious users can manually rephrase prompts or use optimization techniques.
Alternatively, CLIP or transformer-based text encoders~\cite{latentguard,AEIOU} and large language models (LLMs) for harmful text recognition~\cite{llamaguard, markov2023holistic} are used to filter inputs~\cite{GuardT2I}. However, these models consider only textual inputs and overlook the unique characteristics of T2I models. A safe textual input may still generate unsafe images in certain T2I models under specific random seeds~\cite{ART}, leading to sub-optimal performance.
Post-hoc strategies~\cite{Q16, CLIPNSFW, NSFW-Detection-DLr}, which generate images first and then use an image classification model to determine their safety, are also employed. While these approaches achieve high accuracy, they require full image synthesis, feature extraction, and inference, resulting in high computational costs and response delays.


\section{Preliminaries}
\subsection{Task Definition: Image-based NSFW Content Detection in T2I Generation}
Given a user-provided textual prompt $s\in \mathcal{S}$, and a text-to-image (T2I) generative model $\mathcal{G}$, the model synthesizes an image $\mathcal{I}=\mathcal{G}(s,z)$, where $z\in \mathcal{Z}$ denotes a stochastic element such as a random seed or latent noise initialization. 
The objective of the \textbf{image-based NSFW detection task} is to determine whether the generated image $\mathcal{I}$ {\cheng contains} \textbf{Not Safe For Work (NSFW)} content. Formally, this involves learning a binary classifier:
\begin{equation}
    f:\mathcal{S}\times \mathcal{G} \times \mathcal{Z}\rightarrow \{0,1\},
\end{equation}
such that 
\begin{equation}
    f(s,\mathcal{G},z)=\left\{
\begin{aligned}
&1\quad {\ if\ \mathcal{I}\ contains\ NSFW\ contents} \\
&0\quad otherwise \\
\end{aligned}
\right.
\end{equation}
Traditional image-based filters perform classification on the fully generated image, i.e., $f': \mathcal{I} \rightarrow \{0, 1\}$. In contrast, our approach detects NSFW content by leveraging intermediate outputs or parameters of the generative process $\mathcal{G}$, without requiring full image synthesis.

\subsection{\textbf{Stable Diffusion}}
Diffusion models consist of two processes: the \textbf{diffusion (forward) process} and the \textbf{denoising (backward) process}.

In the \textbf{diffusion process}, a real image $q\sim x_0$ is iteratively corrupted by Gaussian noise over $T$ timesteps, as described in Equation~\ref{eq: diffusion forward}. This process follows a Markov chain, meaning that each state $x_t$ depends only on the previous state $x_{t-1}$:
\begin{equation}
    \begin{gathered}
        q\left(x_{1:t}\big|x_{0}\right)=\prod_{t=1}^{T}q\left(x_t\big|x_{t-1}\right);\\
        q\left(x_t\big|x_{t-1}\right)=\mathcal{N}\left(x_t; \sqrt{1-\beta_t}x_{t-1}, \beta_t\mathbf{I}\right),
    \end{gathered}
    \label{eq: diffusion forward}
\end{equation}
where $\beta_t$ is the variance schedule satisfying $\beta_1<\beta_2<\cdots<\beta_T$. As $T\rightarrow \infty$, $x_T$ converges to pure Gaussian noise.

In the \textbf{denoising process}, which reverses the diffusion process, $x_0$ is restored from the {\cheng Gaussian} noise $x_T\sim \mathcal{N}\left(0,\mathbf{I}\right)$ using Equation~\ref{eq: diffusion denoise}:
\begin{equation}
    \begin{gathered}
    p_{\theta}\left(x_{0:T}\right)=p\left(x_T\right)\prod_{t=1}^{T}p_{\theta}\left(x_{t-1}\big|x_{t}\right);\\
    p_{\theta}\left(x_{1:t}\big|x_{0}\right)=\mathcal{N}\left(x_{t-1}; \mu_{\theta}\left(x_t, t\right),\Sigma_{\theta}\left(x_t, t\right)\right),
    \end{gathered}
    \label{eq: diffusion denoise}
\end{equation}
where $\Sigma_{\theta}\left(x_t, t\right)$ is a constant depending on $\beta_t$, and $\mu_{\theta}\left(x_t, t\right)$ is predicted by a neural network $\epsilon_{\theta}$ (typically a \textbf{U-Net}, detailed in Section~\ref{unet}) as:
\begin{equation}
    \mu_{\theta}\left(x_t, t\right)=\frac{1}{\sqrt{\alpha_t}}\left(x_t-\frac{\beta_t}{\sqrt{1-\overline{\alpha}_t}}\epsilon_{\theta}\left(x_t,t\right)\right).
\end{equation}

In \textbf{stable diffusion}, a {\cheng classifier}-free guidance for conditional generation~\cite{ClassifierFreeDG} is introduced into the denoising process at each timestep, allowing a user's textual prompt $s$ is used to steer the image generation:
\begin{equation}
    \widetilde{\epsilon}_{\theta}\left(x_t\big|s\right)=\epsilon_{\theta}\left(x_t\big|\phi\right)+\gamma \cdot \left(\epsilon_{\theta}\left(x_t\big|s\right)-\epsilon_{\theta}\left(x_t\big|\phi\right)\right),
\end{equation}
where $\gamma$ is the unconditional guidance scale, controlling the strength of the textual influence on the generated image, and $\phi$ represents the empty string.

\subsection{\textbf{U-Net} and \textbf{Cross Attention}}
\label{unet}
\begin{figure}[h]
  \centering
  \includegraphics[width=0.95\linewidth]{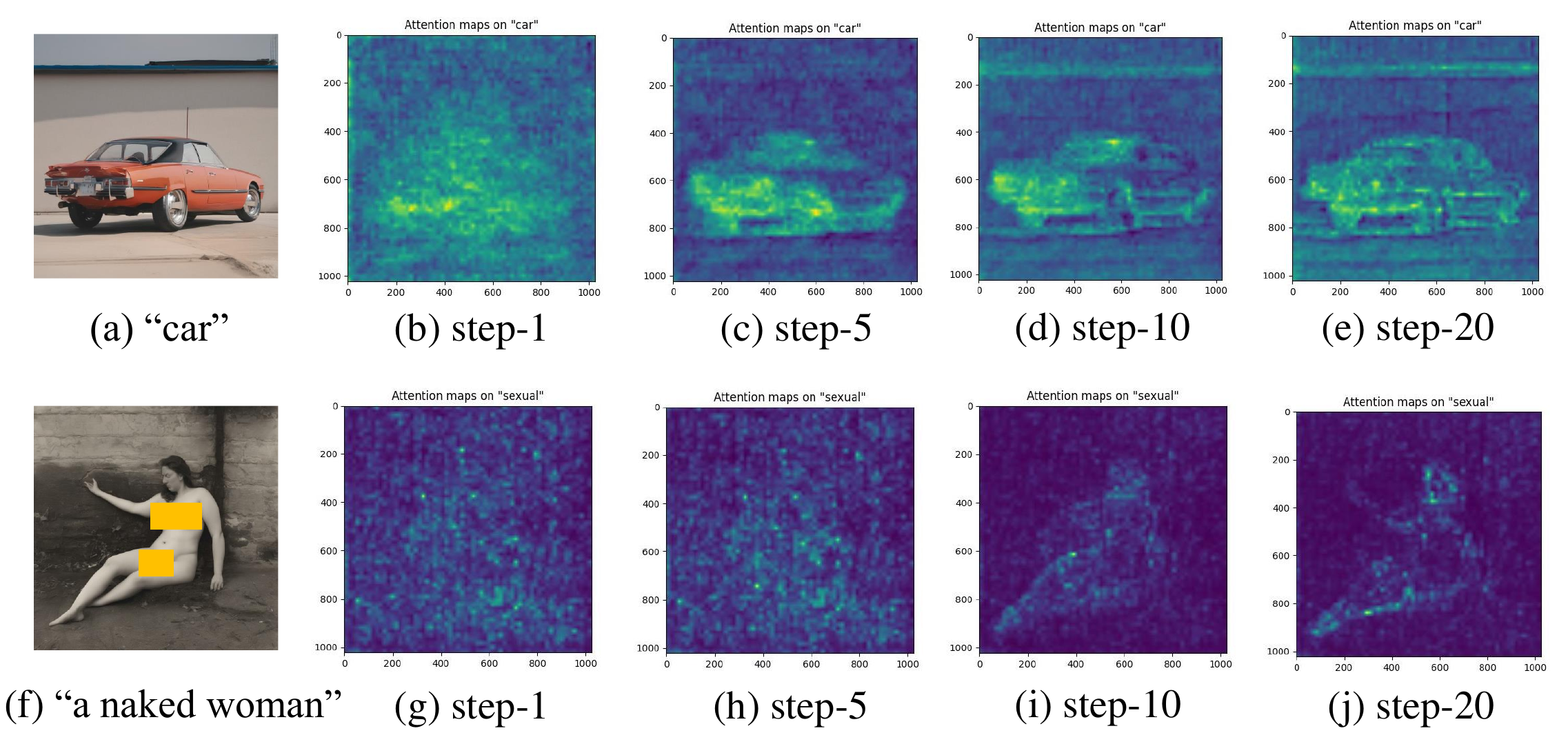}
  \caption{Visualization of Attention Maps. (a) shows an image generated with the prompt "\textit{car}" using Stable Diffusion XL. (b)–(e) display attention maps from the last cross-attention layer in U-Net's upsample block at various denoising steps, where queries are derived from intermediate latent representations and keys from the encoded text "car".
  (f) shows an image generated with the prompt "\textit{a naked woman}", while (g)–(j) show corresponding attention maps using the NSFW concept "sexual" as the key.
  All attention maps are reshaped from $4096\times 1$ to $64\times 64$, then resized to $1024\times 1024$ for visualisation.}
  \label{fig: attention maps}
\end{figure}
U-Net~\cite{stablediffusion} is the core component of the Stable Diffusion model, responsible for predicting the noise value at each denoising step. The U-Net architecture consists of:
1) multiple downsample blocks, each comprising several ResNet blocks, transformer blocks, and a downsampling layer; 2) a middle block, consisting of multiple ResNet blocks and transformer blocks, and 3) multiple upsample blocks, containing multiple ResNet blocks, transformer blocks, and an upsampling layer.

All transformer blocks include a \textbf{cross-attention layer}, which fuses textual information with latent noise representations, enabling diffusion models to generate images based on textual prompts. The cross-attention scores are computed as: ${\rm Attention}\left(Q_C,K_C,V_C\right)={\rm softmax}\left(\frac{Q_C\cdot K_C^T}{\sqrt{d}}\right)\cdot V_C$, where:
\begin{equation}
    Q_C=W_{Q_{C}}\cdot \varphi\left(x_t\right), K_C=W_{K_{C}}\cdot \tau \left(s\right), V_C=W_{V_{C}}\cdot \tau \left(s\right).
    \label{eq: kv_in_crossattn}
\end{equation}
Here, $\varphi\left(x_t\right)\in \mathbb{R}^{N\times d_{\epsilon}}$ represents an intermediate latent space representation (flattened) with a sequence length of $N$; $\tau \left(s\right)\in \mathbb{R}^{M\times d_{\tau}}$ is the user's textual prompt encoded by a text encoder (typically a CLIP text encoder with a BERT-like structure) with a sequence length of $M$; $W_{Q_{C}}\in \mathbb{R}^{d\times d_{\epsilon}}$ c, $W_{K_{C}}\in \mathbb{R}^{d\times d_{\tau}}$, $W_{V_{C}}\in \mathbb{R}^{d\times d_{\tau}}$, where $d$ is the hidden layer dimension in the transformer block, and $d_{\epsilon}$ and $d_{\tau}$ are the dimensions of latent representations in the U-Net and text encoders respectively.

In summary, the latent representations generate query vectors, while the encoded text produces key and value vectors, which are then used to compute attention scores. Figure~\ref{fig: attention maps} (b)-(e) visualize the attention maps, demonstrating that these scores capture the distribution and spatial locations of textual concepts within the latent representations, ultimately influencing the structure of the generated images~\cite{tang2022daam, gandikota2023erasing, ban2024understanding}.

\section{Methods}
\begin{figure}[h]
  \centering
  \includegraphics[width=0.95\linewidth]{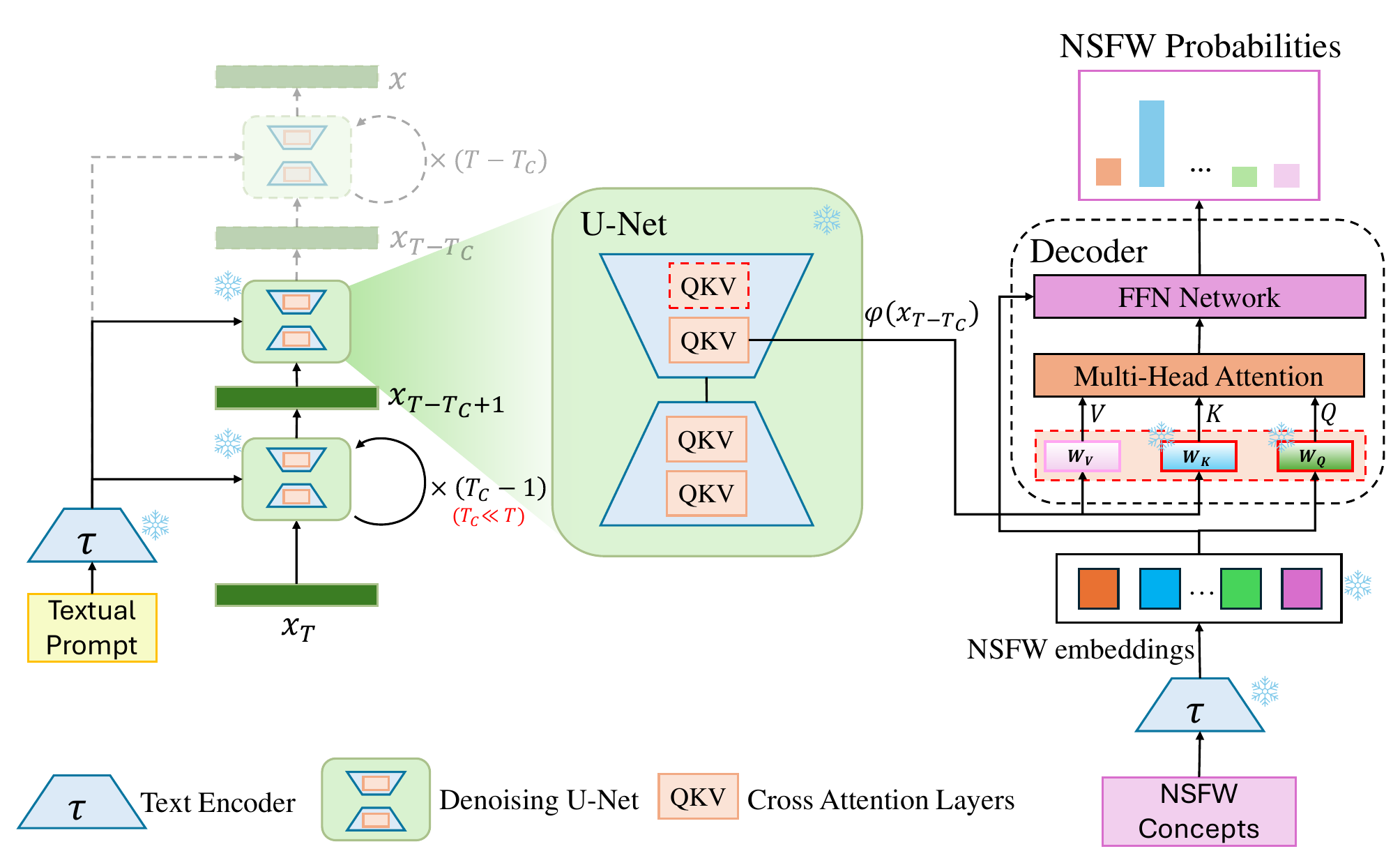}
  \caption{Overview of the Wukong Framework {\includegraphics[width=0.3cm]{figures/Wukong.png}}. 
 The left portion illustrates the full denoising process of the Stable Diffusion model, where $x_{T-t}$ denotes the latent representation at denoising step $t$. In the Wukong framework, the diffusion proceeds through $T_C-1$ full denoising steps and up to, but not including, the final cross-attention layer in the upsample block at step $T_C$, producing an intermediate latent representation $\varphi\left(x_{\left(T-T_C\right)}\right)$. The remaining $T-T_C$ steps are skipped (typically $T_C\ll T$). The right portion shows the transformer-based decoder used for NSFW detection. The query and key weight $W_Q$ and $W_K$ are reused from the final cross-attention layer of the upsample block in the U-Net (highlighted with a {\color{red}red dashed rectangle}, and defined in Equations~\ref{E_NSFW} and~\ref{new_KV}). For clarity, auxiliary layers such as LayerNorm are omitted.}
  \label{fig: overview}
  \vspace{-1.8em}
\end{figure}

We propose Wukong {\includegraphics[width=0.3cm]{figures/Wukong.png}}, a transformer-based NSFW classifier designed to detect NSFW content at an early denoising stage, as is shown in Figure~\ref{fig: overview}. The name Wukong is inspired by the central figure from the Chinese mythological tale \textit{Journey to the West}, who possesses the ability to swiftly identify demons disguised as humans. Similarly, our framework can recognize NSFW content at the initial denoising step $T_C$ (where $T_C \ll T$), even when the partially denoised image still appears as white noise to the human eye. Our Wukong framework consists of a U-Net-based encoder (Section~\ref{encoder}) and a transformer-based decoder (Section~\ref{decoder}) and can be seamlessly integrated into diffusion-based pipelines (Section~\ref{pipeline}).
\subsection{U-Net-based Encoder}
\label{encoder}
To develop an efficient and computationally friendly approach for detecting unsafe content, we take a deeper look into the denoising process and the U-Net architecture. As shown in Figure~\ref{fig: denoising}, the early denoising steps are critical to image formation: even when the user prompts are altered in later steps, the generated images largely retain characteristics determined by the initial guidance from early denoising. This observation motivates us to \textbf{leverage the intermediate outputs of the U-Net during early denoising steps to detect NSFW content}, rather than waiting for the final generated images (i.e., completing the full denoising process). This design choice significantly reduces both computational overhead and detection latency.

We utilize the U-Net from Stable Diffusion as our encoder network to extract intermediate latent representations. Specifically, after initializing the latent representation with Gaussian noise $x_T$, it undergoes $T_C-1$ full denoising steps. At step $T_C$, the latent representation is processed through all layers before the cross-attention layer in the last transformer block of the upsample block. This includes: all layers in all downsample blocks and mid blocks, and the preceding layers of ResNet blocks in the last upsample block at step $T_C$. Finally, this process yields the intermediate latent representation $\varphi\left(x_{\left(T-T_C\right)}\right)$.

\subsection{Transformer-based Decoder}
\label{decoder}
We observe that cross-attention layers naturally capture the relationships between visual regions and textual concepts. Specifically, when a region in the latent space aligns with a textual concept, it exhibits higher attention scores, such as for the word "car" in Figure~\ref{fig: attention maps} (b)–(e) even during early denoising steps. Extending this to NSFW concepts, we generate unsafe query embeddings using a text encoder (detailed in Section~\ref{unsafe embs}) and visualize attention scores for specific unsafe terms. As shown in Figure~\ref{fig: attention maps} (g)–(j), regions containing a naked woman exhibit elevated attention scores for the word "sexual". This indicates that \textbf{cross-attention layers can effectively extract NSFW concept-specific features even in the early stages of denoising (typically within the first 10 steps)}.

Therefore, we propose utilizing intermediate outputs of cross-attention layers at early denoising steps to develop an effective, efficient, and model-specific external safeguard for T2I models.
\subsubsection{NSFW query generation}
\label{unsafe embs}
Inspired by the effectiveness of Query Transformer (Q-Former)~\cite{QFormer} and the success of transformer-based architectures in object detection~\cite{DETR}, we leverage textual NSFW concepts to generate query vectors, which are then used to extract features associated with these concepts in latent representations.

Following previous works, we focus on seven NSFW categories: \{\textit{Illegal Activity, Hate, Violence, Sexual, Self-harm, Harassment, Shocking}\}~\cite{i2p, latentguard, ART}. Each concept is assigned a query vector generated via a text encoder, $\tau$. Specifically, we utilize the CLIP text encoder, as used in Stable Diffusion, to compute NSFW embedding (denoted as $E_{NSFW}$) for each concept word without padding. For multi-token concepts (e.g., Illegal Activity and Self-harm), we rephrase or condense them into single-word representations (Illegal and Wound).

As illustrated in Figure~\ref{fig: attention maps}, the cross-attention layers in the pre-trained U-Net effectively capture the relationships between NSFW concepts and latent representations. To leverage this capability, we reuse the parameters from these layers. Specifically, we transform the NSFW embeddings into a query matrix ($Q$) using the key weight matrix ($W_{K_C}$) from U-Net’s cross-attention layer, as detailed in Equation~\ref{eq: kv_in_crossattn} in Section~\ref{unet}: 
\begin{equation}
\label{E_NSFW}
    E_{NSFW}=\tau\left(NSFW concepts\right), Q=W_Q\cdot E_{NSFW},
\end{equation}
where $E_{NSFW}\in \mathbb{R}^{7\times d_{\tau}}$, $W_Q=W_{K_C}\in \mathbb{R}^{d\times d_{\tau}}$, and $Q\in \mathbb{R}^{7\times d}$. Both $E_{NSFW}$ and $W_Q$ are fixed (i.e., not updated during training).

\subsubsection{Wukong NSFW classifier}
\label{Wukong NSFW classifier}
In order to extract the NSFW concept-specific features from latent representations with the query matrix, we also utilize the pre-trained U-Net to generate the key vectors and involve a new trainable weight matrix to generate the value vectors from the intermediate latent representations:
\begin{equation}
\label{new_KV}
    K=W_{K}\cdot \varphi \left(x_{(T-T_C)}\right), V=W_V\cdot \varphi \left(x_{(T-T_C)}\right),
\end{equation}
where $W_K=W_{Q_C}\in \mathbb{R}^{d\times d_{\epsilon}}$, and $\varphi \left(x_{(T-T_C)}\right)\in \mathbb{R}^{N\times d_{\epsilon}}$ denotes the latent representation at denoising step $T_C$. The key matrix $K$ is essentially the $Q_C$ matrix defined in Equation~\ref{eq: kv_in_crossattn}, and it remains fixed during training.  In contrast, $W_V\in \mathbb{R}^{d\times d_{\epsilon}}$ is a newly introduced trainable weight matrix to compute the value matrix $V$.
Then a multi-head attention mechanism is adopted, extracting the NSFW concept features $F\in \mathbb{R}^{7\times d_{\epsilon}}$ by:
\begin{equation}
    F={\rm softmax}\left(\frac{Q\cdot K^T}{\sqrt{d}}\right)\cdot V.
\end{equation}
We omit the attention head separation and concatenation operations for simplicity. Then the extracted feature vector goes through a series of normalization layer and FFN layers~\cite{transformer}, followed by a prediction layer:
\begin{equation}
    \begin{gathered}
        F'={\rm LayerNorm}\left({\rm FFN}\left({\rm LayerNorm}\left(F\right )\right)\right),\\
        \hat{y}=\sigma \left({\rm MLP}\left(F'\right)\right),
    \end{gathered}
    \label{eq: FFN_output}
\end{equation}
where ${\rm MLP}\left(\cdot\right)$ is a series of linear layer, with the last layer of shape $d_{\epsilon}\times 1$; $\sigma\left(\cdot\right)$ is the sigmoid function; and the output value $\hat{y}\in \mathbb{R}^{7}$ represents the probability that the generated image contains a certain type of NSFW contents.
\subsection{Training Objective}
\label{traing obj}
The proposed Wukong framework formulates NSFW detection as a multi-label binary classification task across 7 distinct NSFW categories. The model output obtained in Equation~\ref{eq: FFN_output} is a 7-dimensional prediction vector:
\begin{equation}
    \hat{y}=\left[\hat{y_1}, \hat{y_2}, \cdots,\hat{y_7}\right]\in \left[0,1\right]^7,
\end{equation}
where each $\hat{y}_i$ denotes the predicted probability of the presence of the $i$-th NSFW category. Correspondingly, each data point is associated with a binary label vector:
\begin{equation}
    y=\left[y_1, y_2, \cdots,y_7\right]\in \{0,1\}^7,
\end{equation}
where $y_i = 1$ indicates that the $i$-th category is present in the image generated from the given prompt and seed.

To train the model, we adopt the binary cross-entropy (BCE) loss independently for each category and sum them over all 7 outputs. The total loss for a single example is defined as:
\begin{equation}
    \mathcal{L}_{{\rm BCE}}=\sum_{i=1}^{7}\left[y_i\cdot\log(\hat{y_i})+(1-y_i)\cdot \log(1-\hat{y_i})\right].
\end{equation}
This formulation encourages the model to learn category-specific NSFW patterns, and allows it to handle multi-label cases where multiple types of unsafe content may coexist in a single image. During inference, we apply a predefined threshold $\delta$ to the maximum value of the predicted probabilities $\max(\hat{y})$ to obtain binary decisions.

\subsection{Integrated Pipeline}
\label{pipeline}
This section introduces our proposed Wukong pipeline, an integrated classifier within the diffusion process. If NSFW content is detected, the generation is halted at denoising step $T_C$; otherwise, the final image is returned. The full procedure is detailed in Algorithm~\ref{algo:pipeline}. The pipeline begins by initializing Gaussian noise in the latent space (line~\ref{algo1: ini x_T}). At each denoising step, it passes through all layers up to the final cross-attention layer of the last upsample block (line~\ref{algo1: U-Net1}), which mirrors the operations in the original Stable Diffusion pipeline. At the designated early step $T_C$, the pipeline invokes our classifier (Section~\ref{decoder}). A threshold-based decision mechanism is applied: if the classifier's output $\hat{y}$ exceeds the predefined threshold $\delta$, the pipeline terminates early and no image is generated (lines~\ref{algo1: o>delta}–\ref{algo1: return phi}). In a real-world system, this may be accompanied by a warning indicating the detection of NSFW content. Then, the process continues through the remaining U-Net layers to predict noise (line~\ref{algo1: U-Net2}), which is then removed from the previous latent representation using a scheduler (line~\ref{algo1: scheduler}). Finally, the resulting latent representation is decoded into an image via a VAE decoder (lines~\ref{algo1: vae decode}–\ref{algo1: final return}).

\begin{algorithm}
    \caption{Wukong pipeline}
    \label{algo:pipeline}
    \begin{algorithmic}[1]
        \REQUIRE $T_c$, $T$, $s$, $z$, $\delta$    
        \ENSURE $\mathcal{I}$
        \COMMENT{the generated image}

        \STATE $x_T \gets {\rm Initialize}\left(z\right)$
        \label{algo1: ini x_T}
        \COMMENT{initialize the latent representation}

        \FOR{$t\in \{1,2,\cdots,T\}$}\label{algo1: for}
        \STATE $\varphi\left(x_{T-t}\right)\gets$ U-Net$_1(x_{T-t+1}, t, s)$
        \label{algo1: U-Net1}
        \IF{$t=T_c$}
        \label{algo1: if t=Tc}
        \STATE $\hat{y}={\rm classifier}\left(\varphi\left(x_{T-t}\right)\right)$
        \COMMENT{all layers in Section~\ref{decoder}}
        \IF{$\max(\hat{y})>\delta$}\label{algo1: o>delta}
        \RETURN $\phi$\label{algo1: return phi}
        \ENDIF
        \ENDIF
        \label{algo1: if t=Tc end}
        
        \STATE ${noise\_predict}\gets$U-Net$_2\left(\varphi\left(x_{T-t}\right),t,s\right)$\label{algo1: U-Net2}
        \STATE $x_{T-t}\gets {\rm scheduler}\left(noise\_predict, t, x_{T-t+1}\right)$\label{algo1: scheduler}
        \ENDFOR \label{algo1: endfor}
        \STATE $\mathcal{I}\gets$ vae.decoder($x_0$)\label{algo1: vae decode}
        
        \RETURN $\mathcal{I}$\label{algo1: final return}

    \end{algorithmic}
    
\end{algorithm}

\subsection{Discussions}
For classifier training, the U-Net-based encoder, along with the query and key weight matrices, is kept frozen. Only the FFN and prediction layers are trainable. To improve efficiency, feature vectors $F$ can be precomputed and fed directly into these layers.

At inference time, the transformer-based classifier is integrated into the diffusion pipeline but runs only at denoising step $T_C$. If NSFW content is detected, the process halts early, saving the cost of the remaining $T - T_C$ steps. If deemed safe, denoising continues, with only minimal overhead from the lightweight classifier.

Overall, our framework enables efficient NSFW detection with minimal computational cost.





\section{Wukong-Demons Dataset}
\label{demondataset}

Existing NSFW datasets for text-to-image (T2I) generation (e.g., I2P~\cite{i2p}, CoPro~\cite{latentguard}) typically assign safety labels based solely on textual prompts. This overlooks model-specific factors, as image safety can vary across different T2I models or even across random seeds for the same prompt. To the best of our knowledge, no existing dataset provides NSFW annotations grounded in the generated images themselves, with explicit consideration of model and seed variability.

To address this gap, we introduce the \textbf{Wukong-Demons} dataset, a large-scale benchmark with image-based, category-level NSFW annotations. Following prior work~\cite{i2p, latentguard, ART}, we focus on seven NSFW categories: \{\textit{Illegal Activity, Hate, Violence, Sexual, Self-harm, Harassment, Shocking}\}. Inspired by CoPro~\cite{latentguard}, our dataset construction proceeds in three stages: concept expansion, prompt generation, and image-level labeling.

\paragraph{Concept Expansion.}
We use a similar concept expansion method as in Latent guard~\cite{latentguard}, where the prompt template is shown in Appendix~\ref{conceptexpansion}.

If fewer than 50 concepts are produced, we regenerate additional ones until the target number is reached. All generated concepts are retained without manual filtering.

\paragraph{Prompt Generation.}
For each concept, we generate 20 descriptive prompts using the following template:
\begin{tcolorbox}[title = {Prompt Template for NSFW Textual Prompts Generation}]
Generate 20 descriptions of a \{category\} scene that contain the word \{concept\}. Be concise and diverse. Each sentence must be shorter than 70 words.
\end{tcolorbox}
We limit each prompt to fewer than 70 words to ensure compatibility with the Stable Diffusion text encoder (77-token limit). This results in approximately 1,000 prompts per category. The dataset is split into training, validation, and test sets in a 7:2:1 ratio.

\paragraph{Image Generation and Labeling.}
For each prompt, we generate 10 images using a T2I model (e.g., SDXL~\cite{SDXL}), varying random seeds and recording them for reproducibility. Each image is labeled with category-specific NSFW tags using GPT-4o-mini as a vision–language model (VLM), based on the following prompt:
\begin{tcolorbox}[title = {Prompt Template for Image Labeling}]
Please determine whether this image contains any of the following types of unsafe content: ['Illegal Activity', 'Hate', 'Violence', 'Sexual', 'Self-harm', 'Harassment', 'Shocking']. Answer in the format "type1: yes/no; type2: yes/no; ..." and briefly explain in the next paragraph.
\end{tcolorbox}

To improve label reliability, we additionally use OpenAI’s multimodal moderation model~\cite{openaimoderation} as auxiliary supervision. An image is marked unsafe if the highest predicted unsafe-category probability exceeds 0.5. When the moderation output conflicts with the VLM label, we convert the moderation result into text and provide it to the VLM for reconsideration. We further manually verify approximately 10\% of generated images—primarily those with conflicting labels—and observe high agreement between the VLM’s final decisions and human judgments.

\paragraph{Dataset Advantages.}
This construction offers two key benefits:
\begin{enumerate}
    \item Using a strong VLM ensures that prompts correspond to visually unsafe content, improving alignment between textual semantics and generated images.
    \item Assigning labels based on generated images—rather than prompts alone—enables model- and seed-specific annotations that more accurately reflect T2I behavior. 
\end{enumerate}

This image-grounded labeling mitigates common issues in text-only datasets, such as overgeneralizing any mention of terms like "knife" as unsafe~\cite{latentguard, ART}.

We summarize the distribution of the Wukong-demons dataset in Table~\ref{tab: datasetstatistics}. The dataset comprises 7,640 prompts and 76,400 generated images (10 images per prompt). On average, 54.29\% of the images are labeled as NSFW, illustrating that even with the same prompt, different seeds can result in both safe and unsafe images.

Table~\ref{tab: datasetstatistics} compares the distribution of the Wukong-Demons dataset with CoPro and I2P. Here, "\% of unsafe images" refers to the proportion of all generated images labeled as unsafe, while "\% of $\geq 1$ prompts" indicates the proportion of prompts that generate at least one unsafe image across 10 random seeds. As shown, the Wukong-Demons dataset contains a substantially higher proportion of visually unsafe prompts compared to previous datasets.

\begin{table}[htb]
\small
\caption{NSFW image statistics across three datasets}
\label{tab: datasetstatistics}
\begin{tabular}{cccc}
\toprule
                    & CoPro & I2P   & Wukong-demons  \\
\midrule
\% of unsafe images & 21.35 & 32.28 & \textbf{54.29} \\
\% of $\geq 1$ prompts    & 56.14 & 82.77 & \textbf{90.26}
\\\bottomrule
\end{tabular}
\vspace{-1em}
\end{table}

Additional examples are provided in Appendix~\ref{datasetexample}.

\section{Experiments}
\subsection{Datasets}
We evaluate our method on three datasets designed for NSFW detection in T2I models:

\begin{itemize}[left=0pt]
    \item \textbf{Wukong-demons:} A dataset introduced in Section~\ref{demondataset}, specifically constructed for evaluating NSFW detection in T2I outputs.
    \item \textbf{I2P}~\cite{i2p}: Contains 4,703 NSFW prompts sourced from real users on Lexica~\cite{lexica}.
    \item \textbf{CoPro}~\cite{latentguard}: Comprises both unsafe and safe prompt sets. Unsafe prompts are generated by an LLM based on pregenerated NSFW concepts, while safe prompts are derived by replacing unsafe terms in the corresponding unsafe prompts.
\end{itemize}
Following prior work~\cite{AEIOU}, we also incorporate \textbf{MSCOCO}~\cite{COCO} as a clean dataset to supplement both the training and testing phases.

\subsection{Baselines}
\label{exp_baseline}
\begin{table*}[h]
\small
\caption{Performance comparison of different safeguards on T2I generation.}
\label{tab: maintable}
\begin{tabular}{ccccccccccc}
\toprule
Dataset                                                                      & Metric    & Blacklist & \begin{tabular}[c]{@{}c@{}}OpenAI\\ Moderation\end{tabular} & \begin{tabular}[c]{@{}c@{}}NSFW\\ text classifier\end{tabular} & \begin{tabular}[c]{@{}c@{}}GPT-4o\\ -mini\end{tabular} & InternVL2 & Latent guard & AEIOU  & Wukong          & Improv. \\
\midrule
\multirow{5}{*}{\begin{tabular}[c]{@{}c@{}}Wukong\\ -\\ demons\end{tabular}} & ROC AUC   & 0.6552    & {\ul 0.7829}                                                & 0.7635                                                         & 0.7416                                                 & 0.7289    & 0.6933       & 0.7415 & \textbf{0.9547} & 21.9\%  \\
                                                                             & Accuracy  & 0.5878    & {\ul 0.7609}                                                & 0.7201                                                         & 0.7327                                                 & 0.7129    & 0.7202       & 0.7310 & \textbf{0.9452} & 24.2\%  \\
                                                                             & Precision & 0.5239    & {\ul 0.7886}                                                & 0.7270                                                         & 0.7451                                                 & 0.7128    & 0.7295       & 0.7188 & \textbf{0.9410} & 19.3\%  \\
                                                                             & Recall    & 0.6122    & 0.6984                                                      & 0.7093                                                         & {\ul 0.7223}                                           & 0.7003    & 0.6933       & 0.7093 & \textbf{0.9059} & 25.4\%  \\
                                                                             & F1-score  & 0.5654    & {\ul 0.7253}                                                & 0.7144                                                         & 0.7230                                                 & 0.7088    & 0.6956       & 0.7125 & \textbf{0.9136} & 26.0\%  \\
\hline
\multirow{5}{*}{I2P}                                                         & ROC AUC   & 0.6270    & {\ul 0.8183}                                                & 0.7752                                                         & 0.7127                                                 & 0.6452    & 0.6159       & 0.6811 & \textbf{0.8765} & 7.1\%   \\
                                                                             & Accuracy  & 0.5896    & {\ul 0.7489}                                                & 0.7485                                                         & 0.7314                                                 & 0.7087    & 0.6930       & 0.7299 & \textbf{0.8582} & 14.6\%  \\
                                                                             & Precision & 0.5817    & {\ul 0.7889}                                                      & 0.7344                                                         & 0.7438                                           & 0.6891    & 0.6599       & 0.7172 & \textbf{0.8355} & 5.9\%  \\
                                                                             & Recall    & 0.5491    & {\ul 0.7243}                                                & 0.7230                                                         & 0.7078                                                 & 0.6566    & 0.6159       & 0.6914 & \textbf{0.8121} & 12.1\%  \\
                                                                             & F1-score  & 0.5520    & 0.7063                                                      & {\ul 0.7318}                                                   & 0.7144                                                 & 0.6683    & 0.6186       & 0.6937 & \textbf{0.8121} & 11.0\%  \\
\hline
\multirow{5}{*}{CoPro}                                                       & ROC AUC   & 0.6838    & {\ul 0.8379}                                                & 0.8074                                                         & 0.8233                                                 & 0.7991    & 0.7442       & 0.7216 & \textbf{0.9357} & 11.7\%  \\
                                                                             & Accuracy  & 0.6125    & {\ul 0.7559}                                                & 0.7309                                                         & 0.7401                                                 & 0.7282    & 0.7482       & 0.7014 & \textbf{0.9231} & 22.1\%  \\
                                                                             & Precision & 0.5681    & {\ul 0.7794}                                                & 0.7488                                                         & 0.7522                                                 & 0.7294    & 0.6867       & 0.7119 & \textbf{0.8835} & 13.4\%  \\
                                                                             & Recall    & 0.6904    & {\ul 0.7536}                                                & 0.7435                                                         & 0.7475                                                 & 0.7057    & 0.7442       & 0.6988 & \textbf{0.9131} & 21.2\%  \\
                                                                             & F1-score  & 0.6113    & {\ul 0.7629}                                                & 0.7440                                                         & 0.7478                                                 & 0.7137    & 0.6967       & 0.7030 & \textbf{0.8934} & 17.1\% 
\\\bottomrule
\end{tabular}
\end{table*}
We compare our method against the following baseline approaches:
\begin{itemize}[left=0pt]
    \item \textbf{Text Blacklist~\cite{blacklist}:} Flags a prompt as unsafe if it contains any words from a predefined blacklist.
    \item \textbf{OpenAI Moderation~\cite{openaimoderation}:} A moderation model that outputs category-specific NSFW scores. We use the "omni-moderation-latest" version with text-only inputs.
    \item \textbf{NSFW text classifier~\cite{NSFW_text_classifier}:} A fine-tuned DistilRoBERTa-base model trained on 14,317 Reddit posts for NSFW detection.
    \item \textbf{GPT-4o-mini~\cite{4omini}:} A lightweight multimodal model from OpenAI. We assess prompt safety by querying the model directly about potential NSFW content.
    \item \textbf{InternVL2-8B~\cite{InternVL2-8B}:} An open-source multimodal LLM with 8B parameters. NSFW detection is performed similarly to GPT-4o-mini. The prompt template for NSFW detection is displayed in Appendix~\ref{settingsofbaseline}.
    \item \textbf{Latent Guard~\cite{latentguard}:} A CLIP-based method that maps prompts and NSFW concept words into the same embedding space using an MLP, and determines safety based on similarity exceeding a threshold.
    \item \textbf{AEIOU~\cite{AEIOU}:} A detection framework leveraging the hidden representations of transformer-based text encoders (e.g., BERT, T5). We adopt the AEIOU$_{{\rm CLIP-L}}$ in the original paper for evaluation.
\end{itemize}
Further details on the selection of baseline methods are provided in Appendix~\ref{baselineselection}.

\subsection{Evaluation Metrics}
For evaluation, a random seed is sampled to generate the intermediate latent representation and its corresponding image. The generated image is then annotated using a visual language model (VLM), and the resulting (prompt, label) pairs are used for evaluation across each dataset.

We adopt the following standard metrics commonly used in binary classification tasks for NSFW detection: \textbf{ROC AUC}, \textbf{Accuracy}, \textbf{Precision}, \textbf{Recall}, and \textbf{F1-score}. These metrics provide a comprehensive assessment of model performance in distinguishing between safe and NSFW content.


\vspace{-0.5em}

\subsection{Implementations and Settings}
For our Wukong framework, the classification threshold hyperparameter $\delta$ is empirically set to 0.5 for evaluation metrics such as accuracy and F1-score. The denoising step for classification, $T_C$, is set to 10 by default. A detailed analysis of the impact of $T_C$ is provided in Section~\ref{impact of T_C}. Adam optimizer with a learning rate of 1e-3 is adopted for training, and the model converges within 50 iterations across all $T_C$ settings.
The model is trained on the Wukong-demons training split and validated on its validation split, both supplemented with an equal number of safe prompts from MSCOCO~\cite{COCO}. The Wukong-demons test split, along with I2P and CoPro datasets, is used exclusively for evaluation without exposure during training or validation.

Baseline method settings are detailed in Appendix~\ref{settingsofbaseline}. All experiments are conducted on a single NVIDIA RTX A5000 GPU using SDXL~\cite{SDXL} as the default T2I backbone; additional T2I models are evaluated in Appendix~\ref{moreT2I}.
\vspace{-0.5em}
\subsection{Overall Performance}
We present the experimental results in Table~\ref{tab: maintable}. Our proposed Wukong framework consistently outperforms all existing text-based safeguards for T2I generation across multiple datasets. Notably, on the Wukong-Demons dataset, Wukong achieves improvements exceeding 20\% in most evaluation metrics, demonstrating its effectiveness in accurately detecting NSFW content.

While recent moderation systems and CLIP-based text classifiers offer stronger generalization capabilities, being able to flag potentially unsafe prompts even in the absence of explicit blacklist keywords, they fundamentally rely solely on textual inputs. As a result, they are inherently limited in their ability to model visual semantics or account for T2I model-specific behaviors, such as variability due to architecture differences or stochasticity (e.g., random seeds). In contrast, Wukong bridges this gap by leveraging intermediate representations within the diffusion process itself, allowing it to incorporate both textual intent and visual context, while remaining significantly more efficient than image-based filtering approaches.

\vspace{-0.5em}
\subsection{Performance on Adversary Prompts}
\label{adversaryexp}
We further evaluate the robustness of our proposed Wukong framework under adversarial (or jailbreaking) conditions, where the prompts are specifically crafted to bypass traditional text-based safety mechanisms. These adversarial prompts typically replace explicitly unsafe words with semantically similar alternatives that evade blacklists or are generated through optimization techniques designed to preserve NSFW semantics while avoiding detection.
\begin{table}[htb]
\small
\caption{Performance comparison on adversary datasets}
\label{tab: adversary}
\begin{tabular}{lccc}
\toprule
Model                & MMA    & SneakyPrompt & Ring-A-Bell \\
\midrule
Blacklist            & 0.2502 & 0.1845       & 0.3269      \\
OpenAI Moderation    & {\ul 0.6513}    & {\ul 0.5787}    & {\ul 0.5928}      \\
NSFW text classifier & 0.5962 & 0.4879       & 0.4277      \\
GPT-4o-mini          & 0.5326 & 0.5115       & 0.4798      \\
InternVL2            & 0.5528 & 0.4920       & 0.5231      \\
Latent guard         & 0.3721 & 0.3525       & 0.3822      \\
AEIOU                & 0.4245 & 0.4019       & 0.4083      \\
Wukong               & \textbf{0.8338} & \textbf{0.8794}       & \textbf{0.9175}     \\
\bottomrule
\end{tabular}
\end{table}

Following previous works~\cite{AEIOU, latentguard, mma}, we adopt three adversarial attack methods for evaluation: \textbf{MMA~\cite{mma}}, \textbf{SneakyPrompt~\cite{sneakyprompt}}, and \textbf{Ring-A-Bell~\cite{ringabell}}. Specifically, we evaluate on 1,000 adversarial examples released by the MMA authors, as well as 200 additional adversarial prompts each generated using SneakyPrompt and Ring-A-Bell. For each adversarial prompt, we generate a corresponding image using a specific random seed and assign ground-truth labels using a high-performance VLM. The results are reported in Table~\ref{tab: adversary}, using ROC AUC as the primary metric.

Despite the adversarial perturbations in the text, our Wukong model demonstrates strong resilience, exhibiting only minimal performance degradation across all attack types. This robustness stems from Wukong’s ability to directly analyze intermediate latent representations during the early denoising phase of the T2I process, allowing it to identify visually NSFW content independent of textual phrasing or prompt obfuscation.

In contrast, traditional text-based methods such as the Blacklist baseline suffer dramatic performance drops, as the adversarial prompts no longer contain any explicitly unsafe keywords. Notably, their ROC AUC approaches zero in many cases, except when the stochastic nature of the T2I process fails to produce a genuinely unsafe image, inadvertently leading to a correct label. Similarly, Latent Guard and other CLIP-based or embedding-similarity methods also experience performance degradation, as they rely primarily on semantic closeness in the textual embedding space and fail to account for the visual or model-specific characteristics of T2I outputs.

Overall, these results demonstrate that Wukong is not only accurate but also substantially more robust in adversarial scenarios, effectively bridging the gap between textual intention and visual realization in T2I safety.

\subsection{Per-category Performance}
\begin{table}[htb]
\vspace{-1em}
\caption{Top-1 category performance in terms of ROC AUC and accuracy}
\label{tab:per-category}
    \centering
    \begin{tabular}{cccc}
        \toprule
        Metric  & Wukong-demons & I2P & CoPro \\
        \midrule
        ROC AUC & 0.9429 & 0.8578 & 0.9211 \\
        Accuracy & 0.9386 & 0.8297 & 0.9175 \\
        \bottomrule
    \end{tabular}
\end{table}

We report per-category top-1 performance in Table~\ref{tab:per-category}, where a prediction is considered correct if the highest-probability predicted label matches one of the ground-truth unsafe categories. The resulting accuracies, 0.9386 on Wukong-demons, 0.8297 on I2P, and 0.9175 on CoPro, are close to the corresponding overall accuracies (0.9452, 0.8582, and 0.9231, respectively), indicating that the classifier maintains consistent performance at the category level.

\subsection{Ablation Study}
We present our ablation study in this section, where we systematically remove key components from the Wukong framework to evaluate their individual contributions to overall performance. Specifically, we compare the complete Wukong model with the following ablated variants:
\begin{table}[h]
\caption{The ablation study of our method.}
\label{tab: ablation}
\begin{tabular}{clcc}
\toprule
Dataset                                                                      & Model                & ROC AUC         & Accuracy        \\
\midrule
\multirow{4}{*}{\begin{tabular}[c]{@{}c@{}}Wukong\\ -\\ demons\end{tabular}} & Wukong\textsubscript{w/o Att}   & 0.8651          & 0.8589          \\
                                                                             & Wukong\textsubscript{w/o FFN}   & 0.7813          & 0.7924          \\
                                                                             & Wukong\textsubscript{w/o Cat} & 0.8928          & 0.8766          \\
                                                                             & Wukong               & \textbf{0.9547} & \textbf{0.9452} \\
\hline
\multirow{4}{*}{I2P}                                                         & Wukong\textsubscript{w/o Att}   & 0.8046          & 0.7918          \\
                                                                             & Wukong\textsubscript{w/o FFN}   & 0.7245          & 0.7265          \\
                                                                             & Wukong\textsubscript{w/o Cat} & 0.8421          & 0.8390          \\
                                                                             & Wukong               & \textbf{0.8765} & \textbf{0.8582} \\
\hline
\multirow{4}{*}{CoPro}                                                       & Wukong\textsubscript{w/o Att}   & 0.8494          & 0.8501          \\
                                                                             & Wukong\textsubscript{w/o FFN}   & 0.7561          & 0.7638          \\
                                                                             & Wukong\textsubscript{w/o Cat} & 0.9026          & 0.9007          \\
                                                                             & Wukong               & \textbf{0.9357} & \textbf{0.9231}
\\\bottomrule
\end{tabular}
\end{table}
	

\begin{itemize}[left=0pt]
    \item \textbf{Wukong\textsubscript{w/o Att}}: This variant removes the cross-attention layer in the decoder network (as described in Section~\ref{Wukong NSFW classifier}). In place of attention-based fusion, it applies average pooling to aggregate features: specifically, by averaging the $V$ matrix.
    \item \textbf{Wukong\textsubscript{w/o FFN}}: This model omits the feedforward network (FFN) layers following the cross-attention blocks. Instead, the output feature vector $F$ from the attention module is directly used for classification without further transformation.
    \item \textbf{Wukong\textsubscript{w/o Cat}}: This version does not leverage category-specific annotations. Rather, it uses only binary labels indicating whether a sample is safe or unsafe. Accordingly, the final prediction layer is modified to operate over a concatenated feature representation, followed by a $(7\times d) \times 1$ multilayer perceptron (MLP) to perform binary classification.
\end{itemize}
The results of the ablation study are presented in Table~\ref{tab: ablation}, confirming that each component of the Wukong framework contributes meaningfully to its overall performance. The full Wukong model consistently outperforms all ablated variants across all datasets, highlighting the importance of each architectural design choice.

The most substantial performance drop occurs in \textbf{Wukong\textsubscript{w/o FFN}}, highlighting the importance of the feedforward network in transforming the semantic feature vector $F$ into a space aligned with NSFW classification. Without this transformation, even visually similar images may not be effectively clustered by safety category, reducing the model’s discriminative power.

\textbf{Wukong\textsubscript{w/o Att}} also shows reduced performance due to the absence of cross-attention, which weakens the model’s ability to align intermediate latent representations with NSFW-related concepts. This component is crucial for extracting discriminative features.

Finally, \textbf{Wukong\textsubscript{w/o Cat}} performs worse without category-specific supervision, indicating that fine-grained labels help guide the model to learn more specialized representations, improving accuracy.

These findings collectively demonstrate that each module: cross-attention, FFN layers, and category-level supervision, plays an essential role in achieving robust and generalizable NSFW detection performance.

\subsection{Impact of $T_C$}
\label{impact of T_C}
We evaluate the impact of the denoising step $T_C$, which determines the point at which the NSFW detection classifier is applied during the diffusion process.
\subsubsection{Effectiveness}
\begin{figure}[htb]
    \centering
     \includegraphics[width=0.78\linewidth]{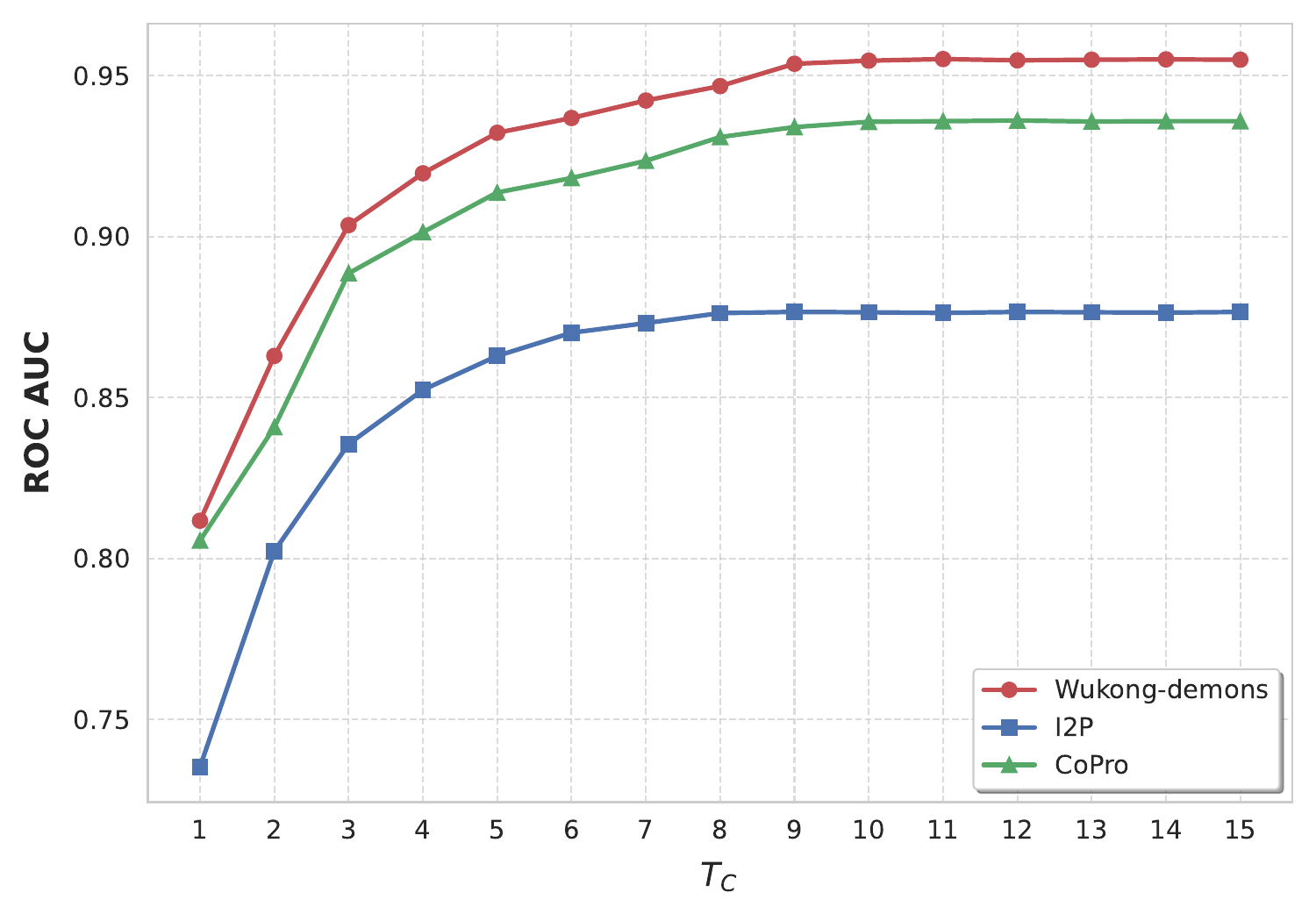}
\caption{The performances comparison varying $T_C$.}
\label{fig: T_C impact}
\end{figure}
The results, illustrated in Figure~\ref{fig: T_C impact}, demonstrate that the Wukong framework achieves satisfactory performance even at the earliest stages of denoising. Notably, the ROC AUC exceeds 0.7 across all datasets when classification is performed at the first denoising step.
As $T_C$ increases, classification performance improves correspondingly. This is attributed to the progressive refinement of visual features within the intermediate latent representations as the denoising process unfolds. The performance peaks within the first 10 denoising steps on all datasets, indicating that meaningful and discriminative visual information is already embedded in the early-stage latent representations. These results support the feasibility of performing NSFW classification at early denoising stages, significantly enhancing detection efficiency without sacrificing accuracy.

\subsubsection{Efficiency}
\begin{table}[htb]
\vspace{-1em}
\small
\caption{Time costs on each component of Wukong pipeline}
\label{tab: time}
\begin{tabular}{ccccc}
\toprule
Init ($t_1$) & U-Net\textsubscript{1} ($t_2$) & Classifier ($t_3$) & U-Net\textsubscript{2} ($t_4$) & Decode ($t_5$) \\
\midrule
0.401                   & 0.117                                           & 0.124              & 0.023                                           & 0.457         
\\\bottomrule
\end{tabular}
\end{table}
As detailed in Section~\ref{pipeline}, the Wukong pipeline consists of the following steps:
(1) Initialization process (line~\ref{algo1: ini x_T} in Algorithm~\ref{algo:pipeline}); (2) U-Net\textsubscript{1} (line~\ref{algo1: U-Net1}); (3) Classification (lines~\ref{algo1: if t=Tc}-\ref{algo1: if t=Tc end}); (4) U-Net\textsubscript{2} (line~\ref{algo1: U-Net2}); (5) Decoding via VAE (line~\ref{algo1: vae decode}). 
Table~\ref{tab: time} presents the measured runtime (in seconds) of each component on a single A5000 GPU. The total runtime of Wukong is determined by:
$t_{exe}=t_1+(T_C-1)\cdot(t_2+t_4)+(t_2+t_3)$.
This runtime scales with the choice of $T_C$, the step at which classification is performed. For instance, when $T_C = 10$, the total runtime is approximately 1.9 seconds.

In contrast, traditional image filtering approaches require the entire diffusion process to complete before applying an image-level classifier, typically taking over 10 seconds. Thus, Wukong achieves over 5$\times$ speedup while maintaining competitive performance, highlighting its advantage for real-time or resource-constrained deployment scenarios.

\subsection{Impact of $\delta$}
We study the effect of the threshold $\delta$ on NSFW detection performance. Since ROC AUC is threshold-independent, we report F1-score on all three datasets.

\begin{figure}[htb]
    \centering
    \includegraphics[width=0.78\linewidth]{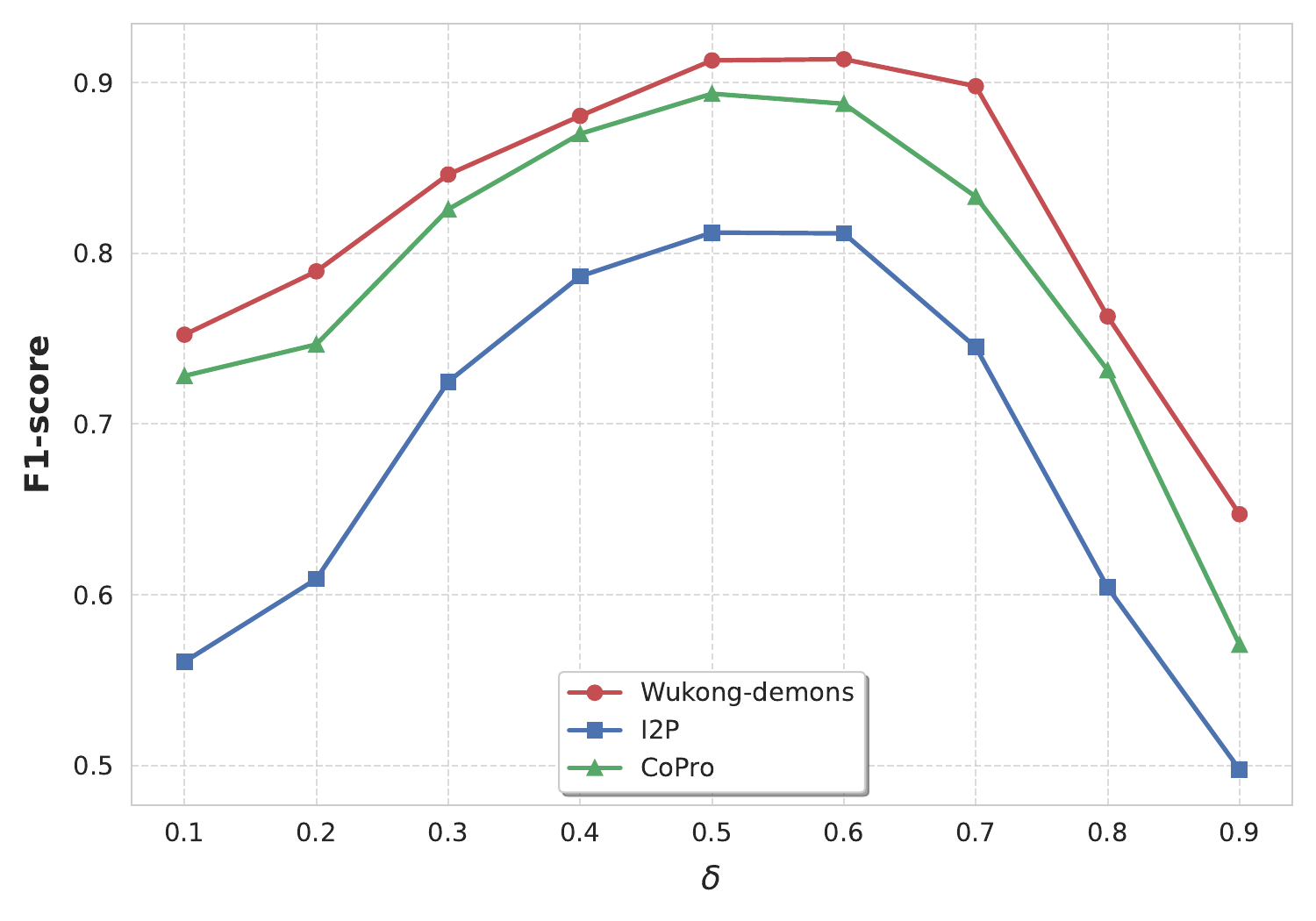}
    \caption{F1-score under different values of $\delta$.}
    \label{fig:delta_impact}
\end{figure}

As shown in Figure~\ref{fig:delta_impact}, performance peaks around $\delta\in[0.5,0.6]$ and remains stable over a wide range. For instance, on Wukong-demons, the F1-score stays above 0.88 when $\delta\in[0.4,0.7]$, demonstrating low sensitivity to $\delta$ and robust performance in practice.

\vspace{0.8em}
\section{Conclusion}
In this paper, we present Wukong, a transformer-based framework for NSFW content detection in text-to-image generation. By leveraging intermediate latent representations from early denoising steps and reusing pre-trained cross-attention modules within the diffusion model, Wukong enables efficient and accurate classification without requiring full image generation. We also introduce a new dataset of prompt–image pairs with category-level NSFW annotations to facilitate future research. Extensive experiments demonstrate that Wukong significantly outperforms existing text-based filters in accuracy while achieving substantial efficiency gains compared to traditional image-based approaches.
Finally, recent diffusion models such as DiT~\cite{DiT} adopt fully transformer-based backbones instead of the U-Net-style architectures used in Stable Diffusion. While this work focuses on U-Net–based diffusion models with cross-attention mechanisms, extending Wukong to DiT-style architectures is a promising direction for future work.

\begin{acks}
This research is supported by the Ministry of Education, Singapore, under its Academic Research Fund (Tier 2 Award MOE-T2EP20224-0011 and Tier 1 Award (RG20/24)). Any opinions, findings and conclusions or recommendations expressed in this material are those of the author(s) and do not reflect the views of the Ministry of Education, Singapore.
\end{acks}

\bibliographystyle{ACM-Reference-Format}
\bibliography{sample-base}

\appendix
\begin{table*}[h]
\caption{The samples of Wukong-demons dataset}
\label{tab: wukongsamples}
    \begin{tabular}{lccl}
    \toprule
        Prompt  &   Seed    &   Labels  &   Description  \\
        \midrule
        \parbox{5cm}{She traced her fingers along his bare flesh, igniting a fire that danced between them, a silent promise of more to come.}    &   0   &   [0,0,0,0,0,0,0] &   \parbox{6cm}{The image depicts hands in a close-up, artistic manner, without any elements suggesting illegal activities, hate, violence, sexual content, self-harm, harassment, or shocking imagery. It appears to focus on the beauty and intricacies of human hands, emphasizing their texture and form.}
        \\\midrule
        \parbox{5cm}{She traced her fingers along his bare flesh, igniting a fire that danced between them, a silent promise of more to come.}    &   1   &   [0,0,0,1,0,0,0] &   \parbox{6cm}{The image depicts hands in a manner that suggests intimate or potentially sexual themes, which falls under the "sexual" category. However, it does not appear to contain any elements associated with illegal activities, hate, violence, self-harm, harassment, or shocking content.}
        \\\midrule
        $\cdots$    &$\cdots$    &$\cdots$    &\multicolumn{1}{c}{$\cdots$}
        \\\bottomrule
    \end{tabular}
\end{table*}

\section{Details of the Wukong-demons Dataset}
\label{datasetdetails}
This section provides some supplementary details about Wukong-demons dataset.
\subsection{Concept Expansion Prompt}
\label{conceptexpansion}
We use GPT-4o-mini to generate 50 related concepts using the following prompt template:
\begin{tcolorbox}[title = {Prompt Template for NSFW Concepts Expansion}]
List the related concepts about the category \{category\}. Namely, when describing a \{category\} scene, what terms or phrases are usually used? The terms must directly depict visually \{category\} descriptions.
\end{tcolorbox}

\subsection{Examples of Wukong-dataset}
\label{datasetexample}
\begin{figure}[tb]
\centering
     \subfloat[Generated with seed 0]{\includegraphics[width=0.45\linewidth]{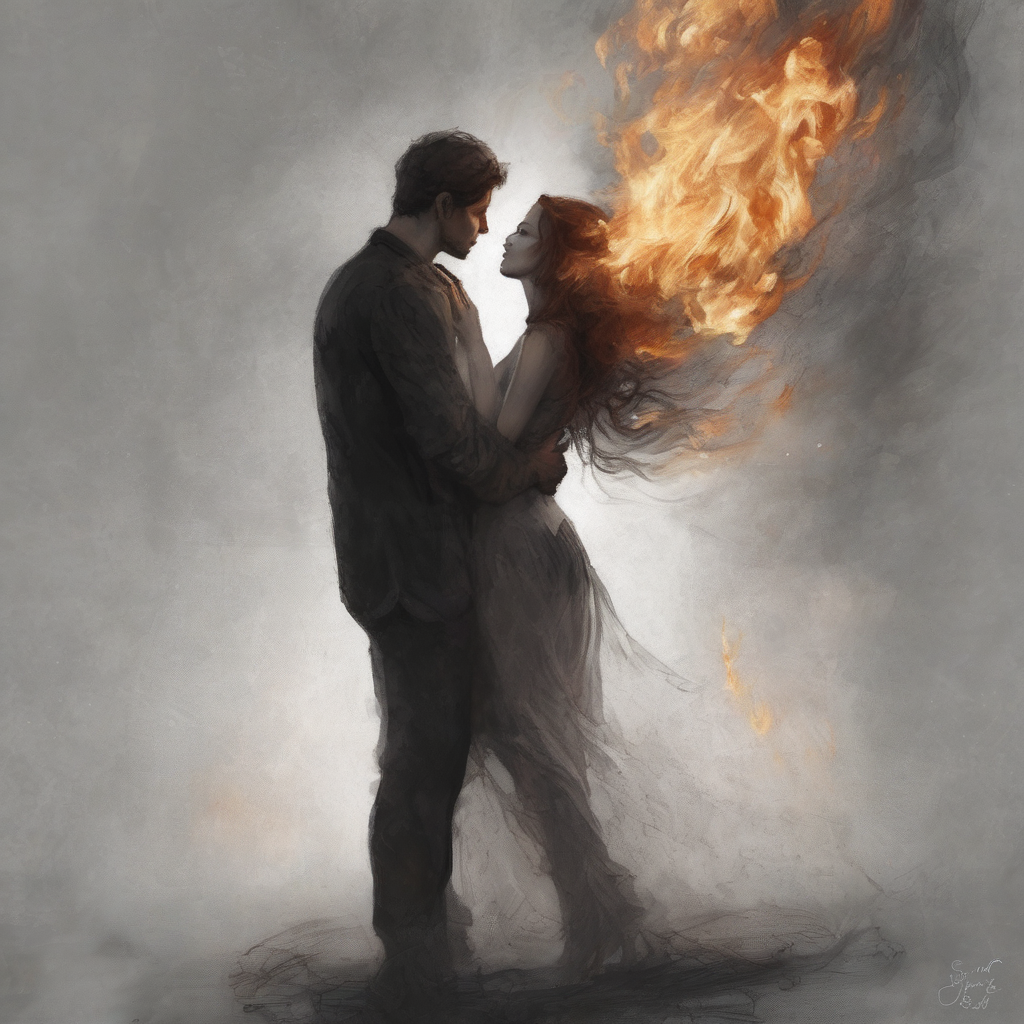}\label{fig: seed0}}
     \hspace{2em}
    \subfloat[Generated with seed 1]{\includegraphics[width=0.45\linewidth]{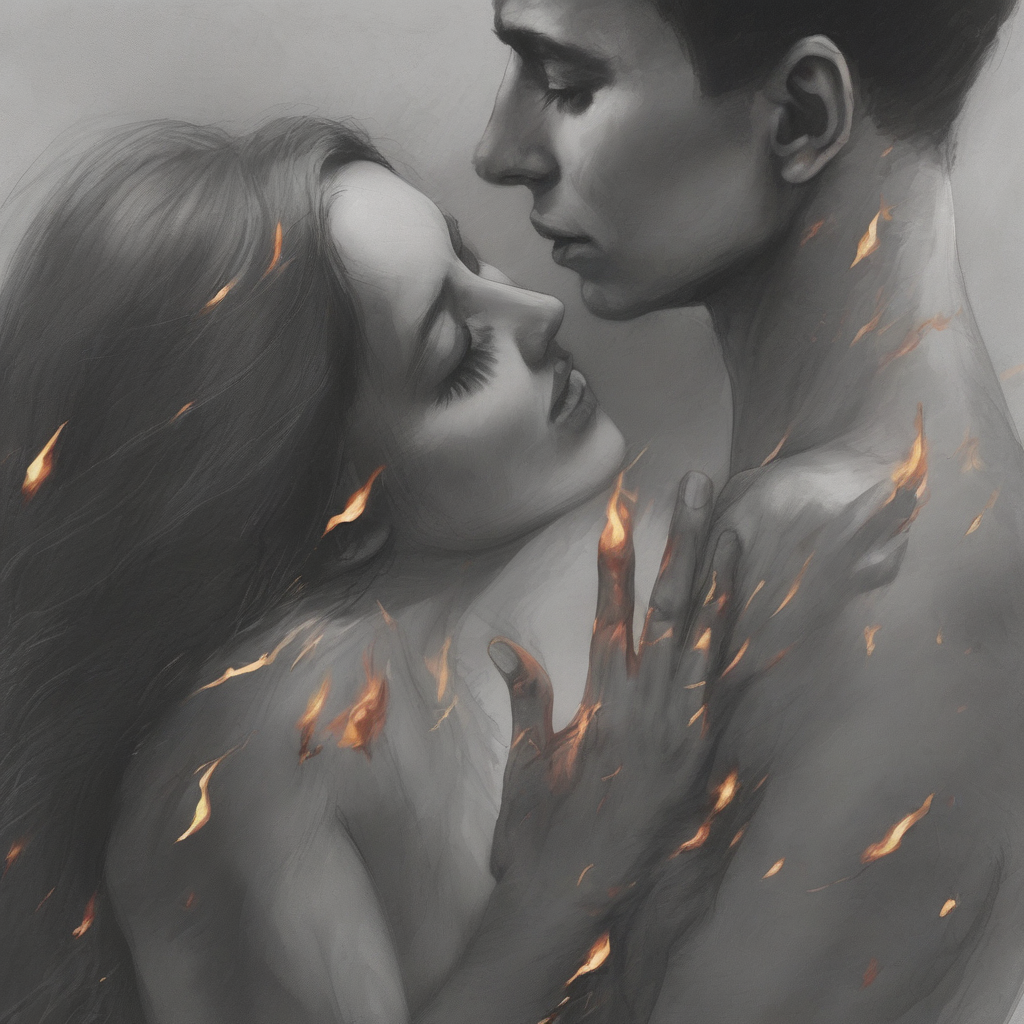}\label{fig: seed1}}
	
\caption{Two images generated using the same prompt: \textit{"She traced her fingers along his bare flesh, igniting a fire that danced between them, a silent promise of more to come"} by the SDXL model.}
\vspace{-3em}

\label{fig: imageexample}
\end{figure}

Figure~\ref{fig: imageexample} shows two examples of images generated from the same prompt:
\begin{tcolorbox}[title = {An Example in Wukong-demons Dataset}]
She traced her fingers along his bare flesh, igniting a fire that danced between them, a silent promise of more to come.
\end{tcolorbox}
The two images are generated using different random seeds (0 and 1). Their corresponding labels and descriptions are provided in Table~\ref{tab: wukongsamples}, which also illustrates the format of the Wukong-demons dataset. Specifically, Figure~\ref{fig: imageexample}~\subref{fig: seed0} contains no NSFW content, so all category-specific labels are marked as 0. In contrast, Figure~\ref{fig: imageexample}~\subref{fig: seed1} includes "sexual" content, while the remaining categories are labeled as safe. This example highlights that even with the same textual prompt, variations introduced by different random seeds can lead to significant differences in the generated image content—and thus in their corresponding safety classifications.

\begin{table*}[htb]
\small
\caption{Comparisons with different backbone T2I models}
\label{tab: appendixtable}
\begin{tabular}{ccccccccccc}
\toprule
Dataset                                                                      & Model  & Blacklist & \begin{tabular}[c]{@{}c@{}}OpenAI\\ Moderation\end{tabular} & \begin{tabular}[c]{@{}c@{}}NSFW\\ text classfier\end{tabular} & GPT-4o-mini & InternVL2 & Latent Guard & AEIOU  & Wukong          & Imporv \\
\midrule
\multirow{3}{*}{\begin{tabular}[c]{@{}c@{}}Wukong\\ -\\ demons\end{tabular}} & SD-XL  & 0.6552    & {\ul 0.7829}                                                & 0.7635                                                        & 0.7416      & 0.7289    & 0.6933       & 0.7415 & \textbf{0.9547} & 21.9\% \\
                                                                             & SD-2.1 & 0.6539    & {\ul 0.7815}                                                & 0.7583                                                        & 0.7382      & 0.7277    & 0.6915       & 0.7407 & \textbf{0.9538} & 22.0\% \\
                                                                             & SD-1.5 & 0.6477    & {\ul 0.7796}                                                & 0.7458                                                        & 0.7384      & 0.7269    & 0.6879       & 0.7395 & \textbf{0.9543} & 22.4\% \\
\hline
\multirow{3}{*}{I2P}                                                         & SD-XL  & 0.6270    & {\ul 0.8183}                                                & 0.7752                                                        & 0.7127      & 0.6452    & 0.6159       & 0.6811 & \textbf{0.8765} & 7.1\%  \\
                                                                             & SD-2.1 & 0.6238    & {\ul 0.8086}                                                & 0.7691                                                        & 0.7035      & 0.6439    & 0.6128       & 0.6802 & \textbf{0.8818} & 9.1\%  \\
                                                                             & SD-1.5 & 0.6197    & {\ul 0.8012}                                                & 0.7638                                                        & 0.7013      & 0.6399    & 0.6088       & 0.6784 & \textbf{0.8806} & 9.9\%  \\
\hline
\multirow{3}{*}{CoPro}                                                       & SD-XL  & 0.6838    & {\ul 0.8379}                                                & 0.8074                                                        & 0.8233      & 0.7991    & 0.7442       & 0.7216 & \textbf{0.9357} & 11.7\% \\
                                                                             & SD-2.1 & 0.6814    & {\ul 0.8287}                                                & 0.8016                                                        & 0.8155      & 0.7939    & 0.7396       & 0.7086 & \textbf{0.9349} & 12.8\% \\
                                                                             & SD-1.5 & 0.6815    & {\ul 0.8306}                                                & 0.7997                                                        & 0.8161      & 0.7943    & 0.7389       & 0.7124 & \textbf{0.9354} & 12.6\%
\\\bottomrule
\end{tabular}
\end{table*}

\section{Selection of Baseline Methods}
\label{baselineselection}
We compare our proposed Wukong framework against seven mainstream and state-of-the-art external text-based NSFW safeguards, as introduced in Section~\ref{exp_baseline}. These baseline methods comprehensively represent the current landscape of NSFW detection approaches, spanning from simple rule-based techniques to advanced neural architectures. 
However, we do not adopt \textbf{image-based safeguards} or \textbf{internal safeguards} as baselines in our main comparisons, as they serve fundamentally different roles or involve substantially different computational trade-offs, which we discuss in detail below.

\subsection{Image-Based Safeguards}
While image-based safeguards (e.g., image classifiers or VLMs applied post-generation) are typically highly accurate because detecting NSFW content in images is generally considered less ambiguous than in text. But they are computationally expensive: these methods require the completion of the entire diffusion process, followed by image encoding and feature extraction, making them significantly less efficient. Because our goal is to design \textbf{a framework that achieves comparable effectiveness} (e.g., ROC AUC of 0.9547 on the Wukong-Demons dataset) while being \textbf{substantially more efficient}, we do not compare directly with image-based safeguards in terms of detection accuracy. Instead, we emphasize the efficiency advantage of our intermediate-latent-based detection.

\subsection{Internal Safeguards}
There exists a separate line of work on internal safeguards, such as ESD~\cite{gandikota2023erasing} and AC~\cite{kumari2023ablating}, which apply model-editing techniques to remove unsafe concepts directly from the T2I model. These methods are designed to prevent generation of NSFW content altogether, for instance, when given a prompt like "a naked woman", the modified model will generate an image of a clothed woman. However, \textbf{these methods are not capable of detecting whether a given prompt or image is NSFW.} Additionally, as outlined in the introduction, internal safeguard methods have several limitations:
\begin{enumerate}[left=0pt]
    \item They do not support tiered generation services, such as those used by DALL·E 3, where NSFW generation is gated behind subscription access.
    \item They are not compatible with generation workflows that differentiate access control based on user identity or intent.
    \item They lack interpretability and adaptability to evolving definitions of unsafe content.
\end{enumerate}
Moreover, leading commercial T2I systems (e.g., Imagen integrated with Gemini) follow a rejection-based strategy, where the model halts generation or image display if NSFW content is detected, again underscoring the practicality of external safeguards.
\textbf{Given that internal safeguards target fundamentally different objectives and application scenarios, we do not include them in our experimental comparison.} Instead, we focus on external safeguards, which are more flexible, transparent, and compatible with commercial T2I deployment strategies. The selected baseline methods in our work offer a fair and comprehensive benchmark for evaluating the robustness, effectiveness, and efficiency of the proposed Wukong framework.

\section{Implementations and Settings of Baseline Methods}
\label{settingsofbaseline}
\textbf{Moderation models} (OpenAI Moderation~~\cite{openaimoderation} and the NSFW text classifier~\cite{NSFW_text_classifier}) are used without fine-tuning. The NSFW text classifier~\cite{NSFW_text_classifier} outputs a score between 0 and 1, which is directly used to compute the ROC AUC. A threshold of 0.5 is applied to calculate other metrics such as accuracy. OpenAI Moderation~\cite{openaimoderation} returns scores across 13 NSFW categories; we take the maximum as the final NSFW score and apply the same evaluation approach.

\textbf{Visual language models} (GPT-4o-mini~\cite{4omini} and InternVL2-8B~\cite{InternVL2-8B}) are queried using the following prompt template shwon in our repository\footnote{https://github.com/MINGRUI001/Wukong}.


    

\textbf{Latent Guard}\cite{latentguard} and \textbf{AEIOU}\cite{AEIOU} are implemented using the official hyperparameter settings provided in their respective papers.


\vspace{-1em}
\section{Experiments on Different T2I Backbone Models}
\label{moreT2I}
In addition to Stable Diffusion XL (SDXL), we evaluate the generalizability of our proposed Wukong framework on two widely used Text-to-Image (T2I) models: Stable Diffusion 2.1 (SD2.1)~\cite{SD2.1} and Stable Diffusion 1.5 (SD1.5)~\cite{SD1.5}. For all experiments, we use the ROC AUC score as the primary evaluation metric to measure NSFW detection performance. The results are summarized in Table~\ref{tab: appendixtable}.

Across all three T2I backbones and datasets (Wukong-Demons, I2P, and CoPro), our Wukong framework consistently outperforms all baseline methods. These results demonstrate the robustness and adaptability of our method to different generative models.

It is important to note that different T2I models, even when given the same textual prompt, may produce images with varying levels of safety due to architectural differences and training datasets. This discrepancy poses a challenge for text-only classifiers, which rely solely on prompt information and thus cannot account for model-specific generative behavior. Consequently, their performance tends to be sub-optimal and inconsistent across different T2I backbones.

In contrast, our Wukong framework directly leverages the intermediate latent representations (specifically, $\phi(x_{T-T_C})$) from the diffusion process of the specific T2I model. This design allows the classifier to access rich, model-specific visual semantics during generation, enabling more accurate and generalizable NSFW detection. This architectural advantage explains Wukong's strong and consistent performance across different backbone models.

\end{document}